\documentclass[letterpaper,conference]{IEEEtran}
\IEEEoverridecommandlockouts

\usepackage{graphicx}
\usepackage{subcaption}
\usepackage{url}   
\usepackage{multirow}
\def\BibTeX{{\rm B\kern-.05em{\sc i\kern-.025em b}\kern-.08em
    T\kern-.1667em\lower.7ex\hbox{E}\kern-.125emX}}
\begin{document}
%\overrideIEEEmargins                                        
% Needed to meet printer requirements.

% In case you encounter the following error:
% Error 1010 The PDF file may be corrupt (unable to open PDF file) OR
% Error 1000 An error occurred while parsing a contents stream. Unable to analyze the PDF file.
% This is a known problem with pdfLaTeX conversion filter. The file cannot be opened with acrobat reader
% Please use one of the alternatives below to circumvent this error by uncommenting one or the other
% \pdfobjcompresslevel=0
% \pdfminorversion=4

% See the \addtolength command later in the file to balance the column lengths
% on the last page of the document

% The following packages can be found on http:\\www.ctan.org
%\usepackage{graphics} % for pdf, bitmapped graphics files
%\usepackage{epsfig} % for postscript graphics files
%\usepackage{mathptmx} % assumes new font selection scheme installed
%\usepackage{times} % assumes new font selection scheme installed
%\usepackage{amsmath} % assumes amsmath package installed
%\usepackage{amssymb}  % assumes amsmath package installed

\title{
Collaborative Robotic Manipulation: A Use Case of Articulated Objects in Three-dimensions with Gravity
%A Use Case of Collaborative Robotic Manipulation: Articulated Objects in Three-dimensions with Gravity
%Collaborative Robotic Manipulation
%A Planning-based Framework for the Manipulation of Articulated Objects in Three-dimensions with Gravity
}

\author{\IEEEauthorblockN{Riccardo Bertolucci\IEEEauthorrefmark{1},
Alessio Capitanelli\IEEEauthorrefmark{2},
Marco Maratea\IEEEauthorrefmark{1}, 
Fulvio Mastrogiovanni\IEEEauthorrefmark{1} and
Mauro Vallati\IEEEauthorrefmark{3}}
\IEEEauthorblockA{\IEEEauthorrefmark{1}Department of Informatics, Bioengineering, Robotics, and Systems Engineering, \\University of Genoa, Genoa, Italy. \\ Emails: \texttt{\{name.surname\}@unige.it}}
\IEEEauthorblockA{\IEEEauthorrefmark{2}Teseo srl, \\Piazza Montano 2a/1, 16126, Genova, Italy.}
\IEEEauthorblockA{\IEEEauthorrefmark{3}School of Computing and Engineering, \\ University of Huddersfield, Huddersfield, United Kingdom. \\ Email: \texttt{m.vallati@hud.ac.uk}}}

%\begin{document}

\maketitle
%\thispagestyle{empty}
%\pagestyle{empty}

%%%%%%%%%%%%%%%%%%%%%%%%%%%%%%%%%%%%%%%%%%%%%%%%%%%%%%%%%%%%%%%%%%%%%%%%%%%%%%%%

\begin{abstract}

This paper addresses two intertwined needs for collaborative robots operating in shop-floor environments.
The first is the ability to perform complex manipulation operations, such as those on articulated or even flexible objects, in a way robust to a high degree of variability in the actions possibly carried out by human operators during collaborative tasks.
The second is encoding in such operations a basic knowledge about physical laws (e.g., gravity), and their effects on the models used by the robot to plan its actions, to generate more robust plans.
We adopt the manipulation in three-dimensional space of articulated objects as an effective use case to ground both needs, and we use a variant of the Planning Domain Definition Language to integrate the planning process with a notion of gravity. 
Different complexity levels in modelling gravity are evaluated, which trade-off model faithfulness and performance.
A thorough validation of the framework is done in simulation using a dual-arm Baxter manipulator. 


\end{abstract}

%%%%%%%%%%%%%%%%%%%%%%%%%%%%%%%%%%%%%%%%%%%%%%%%%%%%%%%%%%%%%%%%%%%%%%%%%%%%%%%%

\section{Introduction}
\label{sec:introduction}

The advent of the Industry 4.0 paradigm is supposed to redefine the nature of shop-floor environments in many directions, such as how robots will be used in the manufacturing process, and the way human operators will interact with them \cite{Krugeretal2009, Heyer2010, Gombolayetal2014}. 
Among the tasks carried out in shop-floor environments, one of the most challenging is related to the manipulation of articulated and flexible objects \cite{HenrichWorn2000, SaadatNan2002}, because in order to achieve a target object configuration it is required to plan a complex sequence of simpler actions, while predicting future object poses as a consequence of those actions, in a full three-dimensional workspace.

The problem of determining the 2D or 3D configuration of articulated (or flexible) objects has originated much research work in the past few years \cite{Wakamatsuetal2006, Agostinietal2011, Nairetal2017, Capitanellietal2017, Capitanellietal2018}, and different strategies have been employed as far as motion planning is concerned \cite{Yamakawaetal2013, Bodenhagen2014, Schulmanetal2016}.
Conceptually speaking, the outcome of such motion planning approaches is a \textit{continuous mapping} in joint space from an initial to a target object's configuration, subject to a number of simplifying hypotheses \cite{Bodenhagen2014}, which lead to two open challenges: 
(i) sequences of manipulation actions are to be robust with respect to errors in perception and execution, as well as to unpredictable actions carried out by human operators in human-robot collaboration (HRC) scenarios, thereby focusing on generalisation and scalability \cite{Agostinietal2011}, and 
(ii) physical laws must be taken into account at the planning level, e.g., the effect of gravity on the object configuration, for predicting its evolution (and therefore estimate) over time \cite{Capitanellietal2017}. 

One reasonable approach to deal with such challenges requires two different (although correlated) representation and reasoning levels:
on the one hand, the collaborative robot needs a \textit{symbolic} representation layer, which can ground planning algorithms able to determine the most appropriate sequence of manipulation actions to attain the target object pose while re-planning when needed, e.g., when human operators decide to intervene \cite{Capitanellietal2017};
on the other hand, those action sequences should take into account such physical laws as gravity, to better predict the expected sequence of intermediate object poses with the aim of assessing the whole manipulation process.
Therefore, two intertwined requirements are defined:
\begin{enumerate}
\item to represent the configurations of complex (articulated or flexible) objects adopting suitable modelling assumptions, and segment the whole manipulation problem in a sequence of actions, each action operating in-between two intermediate 3D object configurations and explicitly considering the effects of exogenous, physical laws, e.g., gravity;  
\item to represent the actions to carry out using a formalism allowing for an easy adaptation to the unpredictable behaviour of human operators.
\end{enumerate}

\noindent The Planning Domain Definition Language (PDDL) represents a possible solution for both requirements, and specifically its hybrid discrete-continuous variant known as PDDL+ \cite{FoxLong2006}. 
PDDL+ enables a clear, predicative-like description of robot operations, while allowing for modelling external processes (and their effects) as they influence the evolution of the phenomenon being modelled, e.g., the effects of gravity on the configuration of an articulated object.
In the past few years, use cases based on PDDL+ have been presented in the literature, e.g., \cite{Crosbyetal2017, Pecoraetal2018, Umbricoetal2018}.
The exploitation of planning techniques for robots has seen a significant growth thanks to the development of such frameworks as ROSPlan \cite{Cashmoreetal2015}, which supports the integration between action and motion planning via predicative knowledge. 
\begin{figure}[t!]
\centering
\includegraphics[width=0.32\textwidth]{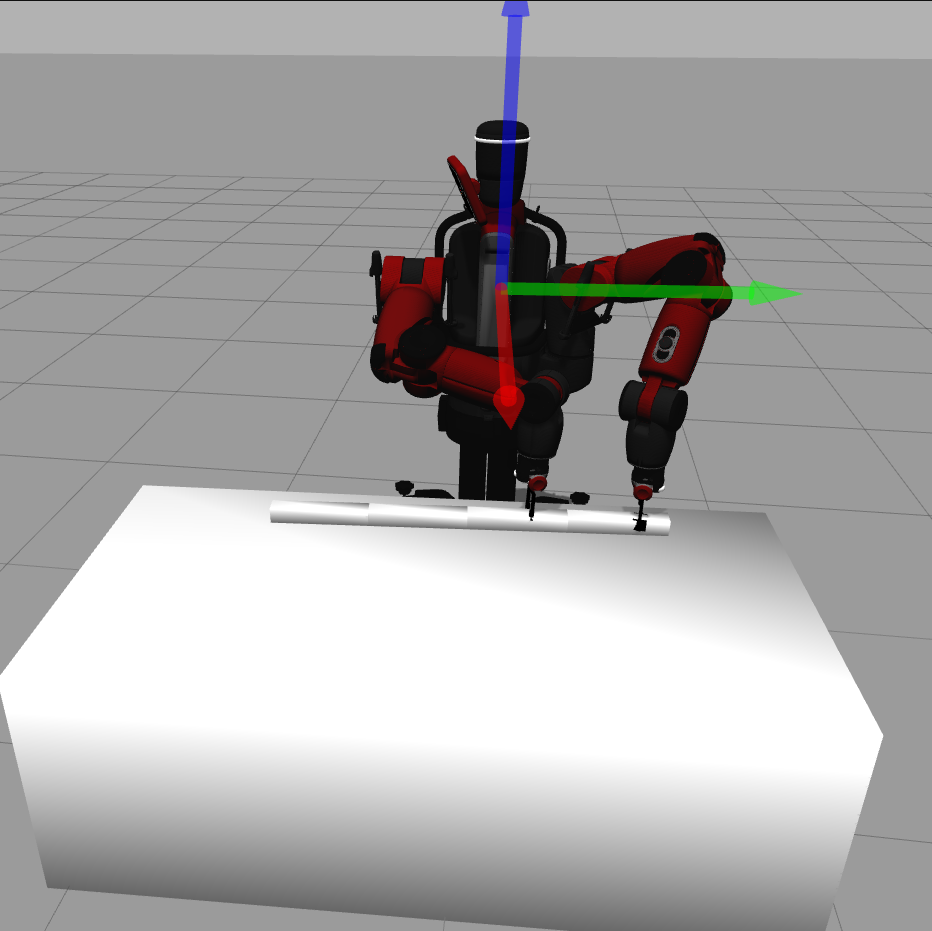}
\caption{Our reference scenario: a simulated dual-arm Baxter manipulator operating on a $4$-link articulated object while taking the effects of gravity into account.}
\label{fig:scenario}
\end{figure}
Based on previous work \cite{Capitanellietal2017, Capitanellietal2018,DBLP:conf/aiia/BertolucciCMMV19}, here we propose a software framework allowing a robot to
(i) plan and execute a sequence of basic actions to manipulate an articulated object, 
(ii) take into account full manipulation in three-dimensional space, including the effects of gravity at different levels of complexity.
We do not explicitly consider here the interaction with human operators, which has been shown elsewhere \cite{Capitanellietal2017, Capitanellietal2018, Darvishetal2018}.
Instead we focus on the architecture-related aspects enabling such interaction.
To this aim, we have setup a scenario where a simulated dual-arm Baxter manipulator operates on an articulated object according to a plan described in PDDL+, which \textit{explicitly} models the effects of gravity (Figure \ref{fig:scenario}). This scenario allows us to validate the overall framework, and to test the suitability of the PDDL+ models in dealing with three-dimensional manipulation. 

The paper is organised as follows.
Section \ref{sec:problem} defines in more detail the problem we aim to solve.
%The overall system's architecture is described in Section \ref{sec:sa}.
The adopted planning formulation is discussed in Section \ref{sec:form}. The architecture and a scenario for performing its validation are described in Section \ref{sec:exp}. 
An experimental analysis of the operationality of PDDL+ models is reported in Section \ref{sec:cpg}.
Discussions and Conclusions follow.

\section{Problem Statement}
\label{sec:problem}

The requirements discussed above lead to a robot perception, representation, planning, and motion architecture characterised by the following behaviours:
(i) similarly to the approach described in \cite{Agostinietal2011}, the robot plans an appropriate series of manipulation actions to determine a sequence of three-dimensional (3D) intermediate configurations for articulated objects\footnote{It is noteworthy that articulated objects are often used as simplified models for flexible objects \cite{Yamakawaetal2013}.} in order to reach a final 3D target configuration;
%the planner tends to avoid \textit{strange} sequences in favour of easy-to-understand plans;
(ii) during plan execution, the robot monitors the outcome of each action, and compares it with the intermediate target configuration to achieve, also taking into account the effects of gravity on each intermediate configuration.

The problem we consider in this paper can be defined as follows:
given a target object configuration in 3D space, determine a plan $\mathcal{P}$ to obtain it as the result of an ordered set of actions:
\begin{equation}
\mathcal{P} = \{a_1, \ldots, a_i, \ldots, a_N; \prec\},
\end{equation}
where each action $a_i$ involves one or more 3D manipulation operations to be executed by a dual-arm robot, subject to the following considerations:
\begin{enumerate}
\item an articulated object is defined in terms of a given number of links and joints, as discussed below;
\item we do not assume an inertial behaviour, i.e., rotating one link causes the movement of all upstream and downstream links, depending on the rotation joint, but subject also to the effects of gravity on all links;
\item we do not postulate any specific grasping or manipulation strategy to obtain a target 3D object configuration starting from another configuration;
\item the perception of articulated objects, although affected by noise, is \textit{perfect}, i.e., data association is given. 
\end{enumerate}

\noindent For the sake of the planning formulation introduced in Section \ref{sec:form}, we define an articulated object as a 2-ple
\begin{equation}
\alpha = \langle \mathcal{L}, \mathcal{J} \rangle,    
\end{equation}
where $\mathcal{L}$ is the ordered set of its $L$ links, i.e.,
\begin{equation}
\mathcal{L} = \{l_1, \ldots, l_j, \ldots, l_L; \prec\},
\end{equation}
and $\mathcal{J}$ is the ordered set of its $J = L-1$ joints, i.e., 
\begin{equation}
\mathcal{J} = \{j_1, \ldots, j_k, \ldots, j_J; \prec\}.
\end{equation}
Each link $l$ is characterised by three parameters, namely a \textit{length}, and two orientations $\theta_l$ and $\gamma_l$, expressed with respect to a robot-centred reference frame (Figure \ref{fig:scenario}).
For computational reasons, we allow only for a limited (parametric) number of discrete orientation values, i.e., $\theta_l$ and $\gamma_l$ can take values from a pre-determined set of possible values.
%in $\mathcal{O}$, where $\mathcal{O} = \{o_1, \ldots, o_{|\mathcal{O}|}\}$.
Given a link $l_j$, we define two sets, namely $up(l_j)$ and $down(l_j)$, such that the former contains upstream links, i.e., from $l_1$ to $l_{j-1}$, whereas the latter includes downstream links from $l_{j+1}$ to $l_L$.
%Orientations are expressed with respect to an \textit{absolute}, robot-centred, reference frame.
Such representation leads to the direct perception of links and their orientations \cite{Capitanellietal2018}.
When a sequence of manipulation actions is planned, changing one absolute orientation requires, in principle, the propagation of such change upstream or downstream the object via joint connections.  
Given an articulated object $\alpha$, its configuration is a L-ple
\begin{equation}
\mathcal{C}_{\alpha}=\{(\theta, \gamma)_1, \ldots, (\theta, \gamma)_l, \ldots, (\theta, \gamma)_L\},
\end{equation}
where it is intended that the orientations $\theta$ and $\gamma$ are expressed with respect to an \textit{absolute}, robot-centred, reference frame.

%\section{System's Architecture}

\section{Formulation of the Planning Problem}
\label{sec:form}

\subsection{General Formulation}

In our architecture we exploit the hybrid discrete-continuous representation capabilities offered by PDDL+ to formulate an explicit encoding of the effects of gravity on the configurations of articulated objects, while they are manipulated.
Specifically, we developed three different domain models, corresponding to three abstraction levels about the role of gravity in the planning process.
%To simulate the impact of gravity on the joint angles $\phi_j$, in our model we represent angles as {\it absolute}: angles originate
%from link orientations expressed with respect to a unique, typically robot-centred, reference frame. 
It is noteworthy that an absolute representation of joint angles, as postulated above, has non obvious consequences on the planning formulation:
on the one hand, it reduces the burden on the robot sensory and perceptual system, because link orientations are directly observable, and do not require any additional computation;
on the other hand, the overall computational complexity of the planning process is increased, due to the fact that manipulation actions must be propagated to all the upstream or downstream link orientations, and later being updated by the effects of gravity.
The interested reader is referred to our previous work discussed in  \cite{Capitanellietal2018} for an extensive comparison of different joint angle representation techniques, although limited in that case to a 2D workspace.

In the PDDL+ models described in this paper\footnote{All developed models and the material used for simulations is available at: https://github.com/Flaudia/AMAO.}, a \texttt{connected} predicate is used to describe the fact that two links are connected by a joint.
It is worth mentioning that joints are not explicitly modelled in the representation.
The \texttt{connected} predicate indicates the presence of a joint between the two involved links, and the related orientation is given via the \texttt{angle} function, which indicates the absolute orientation of a link with respect to one axis.
The values of angles range between $0$ and $359$ degrees (in real-world objects, this range may be reduced).
The effects of the manipulation of two \texttt{connected} links are propagated via a corresponding \texttt{affects} predicate. 
In order to reduce the overall computational complexity associated with the propagation, we fixed \textit{a priori} the way in which the robot can manipulate two connected links, so that the propagation of angle values can only happen upstream.
In other words, given two links, we allow the robot to move only the upstream link, while the other one is kept still by one of the robot grippers.
However, such an assumption does not limit the generality of the models.
If needed, the models can be easily extended to deal with both up and downstream manipulations by adding the appropriate predicates. 
The number of planes whereby angles can be grounded with is not defined \textit{a priori}, and can be easily modified.
In our scenario we consider two planes, i.e., vertical and horizontal, corresponding to a full 3D workspace. 

Link orientations can be modified by a planning engine using the following PDDL+ constructs:
\begin{itemize}
\item
An operator \texttt{start-increase(l1,l2,plane)} is used to modify the orientation of the link \texttt{l2} on the \texttt{plane} by using one gripper to keep \texttt{l1} still, and another gripper to rotate \texttt{l2}. 
\item
A process \texttt{move-increase(l2, plane)} is used for modelling the continuous movement performed by the robot to increase the absolute angle related to \texttt{l2} on the corresponding \texttt{plane}. 
This process is activated by the \texttt{start-increase} operator.
\item
An operator \texttt{stop-increase(l1,l2,plane)} is activated by the planner to stop the change in relative orientation between \texttt{l1} and \texttt{l2}.
The robot is afterwards releasing the two links.
\item
The two related events \texttt{back-to-zero(l,plane)} and \texttt{back-to-360(l,plane)} are triggered when the value of the angle of link \texttt{l} with respect to \texttt{plane} reaches $359$ (respectively, $0$).
In the former (latter) case, the value of the angle is reset to $0$ (respectively, $359$).
\item 
A process \texttt{propagate-increase(l1,l2,plane)} is activated when \texttt{move-increase(l2,plane)} is active.
It enables propagating the effects of the current manipulation on all the affected upstream angles. 
\end{itemize}

\noindent The constructs above are in charge of modelling (and afterwards performing) manipulation actions to modify joint angle values.
A corresponding set of constructs is used to allow the planner to decrease some specified angles. 
Figure \ref{fig:pddl+} shows the PDDL+ encoding of the \texttt{start-increase} operator and the \texttt{back-to-zero} event.
It is noteworthy that the predicate \texttt{in-use} is exploited to avoid parallel manipulations of the articulated object by the robot.
This is because the robot grippers are not explicitly modelled by the encoding, and therefore \textit{many} different actions could potentially be planned in parallel by the planning engine.
The \texttt{free-to-move} predicates are used to indicate whether a link is currently being manipulated or not.
These predicates act as a sort of token for grasping a specific link. 
%The function \texttt{speed-i} represents the speed of the manipulation, for increasing an angle, that can be performed by the considered robot. 

\begin{figure}[t]
%\footnotesize
%\small
\begin{verbatim}
(:action start-increase
:parameters (?l1 -link ?l2 -link ?x -plane)
:precondition (and
    (connected ?l1 ?l2)
    (not (in-use)))
:effect (and
    (in-use)
    (not (free-to-move ?l2))
    (not (free-to-move ?l1))
    (increasing-angle-robot ?l2 ?x)))

(:event back-to-zero
:parameters (?l3 -link ?x -plane)
:precondition 
    (>= (angle ?l3 ?x) 360)
:effect  
    (assign (angle ?l3 ?x) 0))
\end{verbatim}
\caption{Part of the proposed PDDL+ formulation.}
\label{fig:pddl+} 
\end{figure}

%descrizione del dominio PDDL+

\subsection{Modelling Gravity}

In a general perspective, the effects of gravity are
(i) \textit{continuous} in nature (as they can be modelled by physical laws), and
(ii) not under the \textit{direct} control of the planning engine.
It comes without saying that for a collaborative robot to model the effects of gravity while planning manipulation actions can be of the utmost importance for the predictive power of each single action, to be seen in a sensorimotor loop perspective.
%, as well as for increasing its naturalness in the eyes of human operators during collaborative actions.
Such PDDL+ constructs as \textit{continuous processes} and \textit{events} can be extremely useful to accurately describe the effects of gravity on the manipulation of articulated object in a full 3D workspace. 
However, an accurate representation of such effects can be computationally cumbersome, and may prevent the generation of valid plans in a reasonable amount of time, which may jeopardise a prompt reaction to human operator actions.
Because of that, we introduce three different levels of complexity, which can be encoded in the proposed PDDL+ models.
It is noteworthy that the \textit{typical} articulated object, in order to support its manipulation via a robot, has quite stiff joints, which are therefore resistant -- up to a certain degree -- to the effects of gravity.
 
\textit{Level \#1: No Gravity} (NG).
The most trivial way to reduce the complexity burden due to the computation of the effects of gravity on the articulated object is, of course, to completely ignore gravity.
In cases whereby the joints of the articulated object are very stiff, this model can still give some useful information to the robot.
Arguably, the reduced complexity may allow for quickly re-planning when the robot \textit{observes} that gravity has significantly modified the object pose and therefore its configuration.
    
\textit{Level \#2: Uniform Circular Motion} (UCM).
A more sophisticated way for modelling the impact of gravity on an articulated object can be obtained by taking a joint-by-joint perspective.
As links are connected by joints, they cannot simply fall to the ground, but are bounded to each other via joints. 
The effects of gravity on a joint angle can be then modelled as a uniform circular motion that tends to orient the angle towards a value of $360$ (if we consider a $180$-degree angle to be on the $z$ axis).
In this encoding, the angular speed is assumed constant.
The impact of gravity on an angle is modelled using a pair of dedicated processes (according to the fact that the initial angle is lower or higher than $180$) and, since angles are absolute, such motion is also propagated to all the affected joints via a dedicated PDDL+ process. 
The effects of gravity on a joint can be stopped for two reasons:
(i) the angle has reached the rest position ($360$/$0$ degrees), or
(ii) the corresponding link is grasped by the robot gripper.
    
\textit{Level \#3: Uniformly Accelerated Circular Motion} (UACM). 
Building on top of the UCM formalisation, we introduce a more advanced representation of the effects of gravity by modelling it as a uniformly accelerated circular motion.
As done above, all joint angles tend to return to a $360$ degree position on the $z$ axis. 
However, their initial angular speed is $0$, but it is uniformly accelerated.
The acceleration is encoded in PDDL+ using an additional process, which is in charge of increasing the angular speed while the gravity effect is active on a specific joint, and an appropriate event that \textit{resets} the speed value when the effect of gravity ends. 

As far as the change of angle values performed by the robot is concerned, in both the UCM and the UACM formulations the effects of gravity on an angle are propagated to all the affected joints by a set of dedicated processes and events.
An example of the processes exploited in the UCM formulation for modelling gravity is provided in Figure \ref{fig:pddl+2}. 
The process, as the dual \texttt{gravity-decrease} (not shown) and those exploited in the UACM formulation, keep a Boolean predicate true, which is used to represent the fact that gravity is impacting the corresponding link.
Semantically, this implies that every time step when the process is active, the corresponding predicate is set to true.
While this can be possibly interpreted as an abuse of PDDL+ language features, it is supported by some state-of-the-art planning engines.
All encodings are present in the Git repository.

\begin{figure}[t!]
%\small
\begin{verbatim}
(:process gravity-increase
:parameters (?l1 - link)
:precondition (and 
    (free-to-move ?l1)
    (> (angle ?l1 z-axis) 180)
    (< (angle ?l1 z-axis) 360))
:effect (and
    (increase 
        (angle ?l1 zaxis) (* #t (speed-g)))
    (increasing-angle-gravity ?l1)))
\end{verbatim}
\caption{\label{fig:pddl+2} The process used in the UCM formulation to model the effects of gravity on angles ranging between $180$ and $359$ degrees.}
\end{figure}

\section{Architecture}
\label{sec:exp}

This Section describes the scenario and how the architecture has been implemented for the considered case study. One illustrative example is then analysed.
%validating the use of the PDDL+ planning for manipulating objects, and details the results observed on one illustrative example. 

\subsection{Scenario}
\label{sec:scenario}

We have setup a simulation scenario where a dual-arm Baxter manipulator operates on articulated objects to modify their configurations in a full 3D workspace (Figure \ref{fig:scenario}).
Considering the obvious perception and manipulation challenges associated with articulated objects, as well as the need to conduct a significant number of experiments to assess computational performance with different gravity models (i.e., NG, UCM, and UACM) and with a varying number of links, simulation has been preferred over experiments in real-world conditions.
Experiments with a real robot platform are in the works.
Baxter has been selected because it is a widely used research platform, and it is provided with two $7$ DoF arms, which is fundamental for rotating links with respect to each other in three dimensions.
It is noteworthy that our models have been implemented within the \textsc{PlanHRC} architecture, therefore adding PDDL+ compliant capabilities to the previous framework \cite{Capitanellietal2018}.
The techniques described in this paper can be applied to other robot platforms, either in simulation or in real-world conditions, as long as appropriate perception, low-level motion planning algorithms, and manipulation strategies are adopted. 

In our scenario, perception is mimicked using a virtual camera device located on top of the robot's \textit{head} and pointing downward to the robot workspace, which provides a 6D pose for each link. 
Acquired information about link configurations undergoes an estimate process, and the generated predicative knowledge is encoded in an OWL-DL ontology \cite{Krotzschetal2012}. 
When a target object configuration is imposed, a planning process is activated.
Using the models described above and the knowledge currently encoded in the ontology, such process generates action plans, which are then carried out by executing each action via motion planning and execution.
We adopt ENHSP as a planning engine \cite{Scalaetal2016,Ramirezetal2018}, and the well-known MoveIt framework for motion planning and execution.
ENHSP has been selected on the basis of a comparative analysis involving different planners supporting PDDL+ \cite{DBLP:conf/aiia/BertolucciCMMV19}, and of its good performance on PDDL+ benchmarks \cite{DBLP:conf/iccS/FrancoVLM19}.
An extensive analysis of the overall performance of ENHSP is out of the scope for our discussion.
Our architecture runs on the Robot Operating System (ROS, Indigo release) framework using Gazebo 2 as a simulation environment. 
Simulations run on an Ubuntu 14.04 machine, equipped with an Intel Core i7-4790 CPU clocked at 3.60 GHz, and 16 GB of RAM.
As noted above, the simulation material has been made freely available online.

\subsection{One Illustrative Example}
\label{sec:example}

In this Section we briefly discuss one example corresponding to a specific manipulation problem instance. 
Each simulation generates randomly an initial articulated object's configuration, and adds Gaussian noise to joint angle values.
The configuration is perceived by the virtual camera on the robot, each link's pose is estimated, and then the corresponding predicates are generated in the ontology.
A target object configuration is randomly generated as well.
When this happens, PDDL+ problem files are generated on the basis of one of the three different gravity formulations presented above, and then submitted to the planning module.
Although in different simulations we employed articulated objects with a varying number of links (up to $L = 12$) and therefore joints, in this particular case we consider an articulated object with $L = 4$ links (each one approximately $15$ cm long and $3$ cm thick), and $J = 4$ joints.

\begin{figure}[t!]
%\small
\begin{verbatim}
(:init
(= (speed-i) 10)
(= (speed-d) 10)
(= (speed-g) 0.5)
(= (angle L1 xy-axes) 0.0)
(= (angle L1 z-axis) 0.0)
(= (angle L2 xy-axes) 0.0)
(= (angle L2 z-axis) 0.0)
(= (angle L3 xy-axes) 0.0)
(= (angle L3 z-axis) 0.0)
(= (angle L4 xy-axes) 0.0)
(= (angle L4 z-axis) 0.0)
)

(:goal (and
	(> (angle L2 xy-axes) 265.4)
	(> (angle L2 z-axis) 85.5)
	(> (angle L3 xy-axes) 246.8)
	(> (angle L3 z-axis) 65.0)
	(< (angle L4 xy-axes) 33.4)
	(< (angle L4 z-axis) 5.5)
))
\end{verbatim}
\caption{An example of initial and final states where: \texttt{speed-i} and \texttt{speed-d} represent how many degrees per second the robot will increase or decrease the joint angle it is acting on, while \texttt{speed-g} represents how many degrees per second a joint affected by gravity will lose over time if the robot is not acting on it.} 
\label{fig:example_is_fs}
\end{figure}
\begin{figure}[t!]
%\small
\begin{verbatim}
;1.0: (back-to-360 L4 z-axis)
1.00: (start-decrease L1 L2 z-axis)
;8.0: (back-to-360 L3 z-axis)
9.00: (stop-decrease L1 L2 z-axis)
10.0: (start-increase L1 L2 xy-axes)
;11.0: (back-to-zero L4 z-axis)
;11.0: (back-to-zero L3 z-axis)
12.0: (stop-increase L1 L2 xy-axes)
12.0: (start-decrease L3 L4 z-axis)
;13.0: (back-to-360 L4 z-axis)
;13.0: (back-to-360 L2 z-axis)
13.0: (stop-decrease L3 L4 z-axis)
13.0: (start-increase L3 L4 z-axis)
\end{verbatim}
\caption{An excerpt of a sample plan obtained using the the UCM PDDL+ model. The plan includes also events that are triggered during the execution (indicated using a "\texttt{;}"). Processes are removed for readability (the propagation of manipulation effects on angles can lead to a large number of processes to be active at the same time).
%The last line means that the goal configuration is achieved $2$ seconds after the execution of the last action.
}
\label{fig:plan}
\end{figure}
%;1.0: (back-to-360 L4 zaxis)
%1.00: (start_movement_decrease L1 L2 zaxis)
%;3.00: (back-to-360 L5 zaxis)
%;8.0: (back-to-360 L3 zaxis)
%9.00: (stop_movement_decrease L1 L2 zaxis)
%10.0: (start_movement_increase L1 L2 xyaxes)
%;11.0: (back-to-zero L5 zaxis)
%;11.0: (back-to-zero L4 zaxis)
%;11.0: (back-to-zero L3 zaxis)
%12.0: (stop_movement_increase L1 L2 xyaxes)
%12.0: (start_movement_decrease L3 L4 zaxis)
%;13.0: (back-to-360 L5 zaxis)
%;13.0: (back-to-360 L4 zaxis)
%;13.0: (back-to-360 L2 zaxis)
%13.0: (stop_movement_decrease L3 L4 zaxis)
%13.0: (start_movement_increase L3 L4 zaxis)
%;; (wait for 2.0 seconds)

Figure \ref{fig:example_is_fs} shows initial and final states for the example we consider here, whereas a sample plan is shown in Figure \ref{fig:plan}, obtained using the UCM formulation.
It can be noted that each action involves one rotation operation on the target link, specifically around a certain axis.
As the consequence of specific actions, such events as \texttt{back-to-360} and \texttt{back-to-0} are triggered independently from the effect of gravity.
While the plan is executed, such information can be used as a prediction of the expected object configuration in the near future for motion planning and execution. 

\begin{figure}
    \centering
    \begin{subfigure}[b]{0.213\textwidth}
        \includegraphics[width=\textwidth]{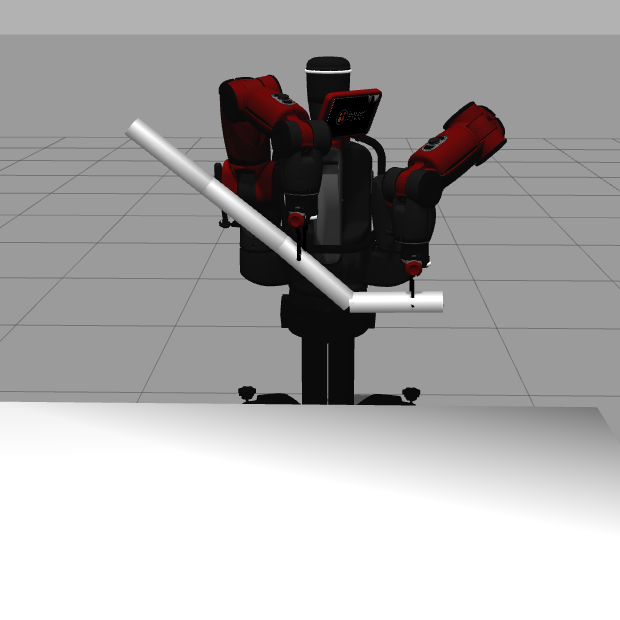}
        \label{fig:step1}
    \end{subfigure}
    \begin{subfigure}[b]{0.213\textwidth}
        \includegraphics[width=\textwidth]{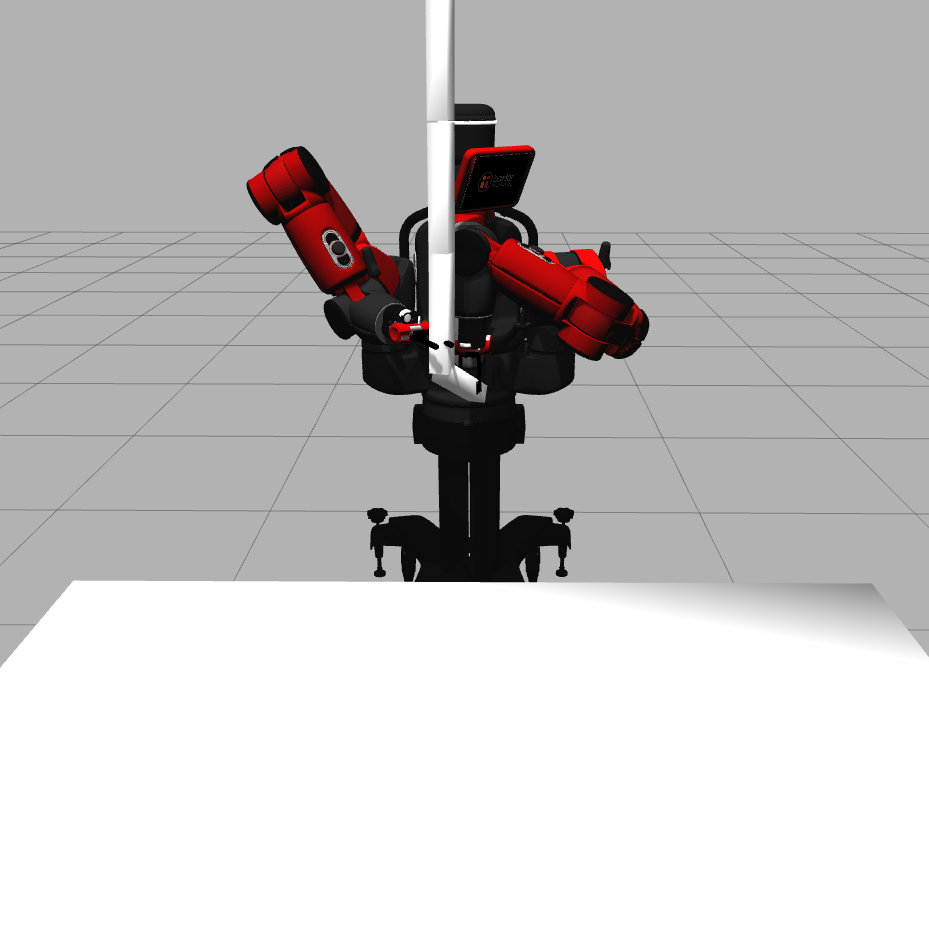}
        \label{fig:step2}
    \end{subfigure}
    \begin{subfigure}[b]{0.213\textwidth}
        \includegraphics[width=\textwidth]{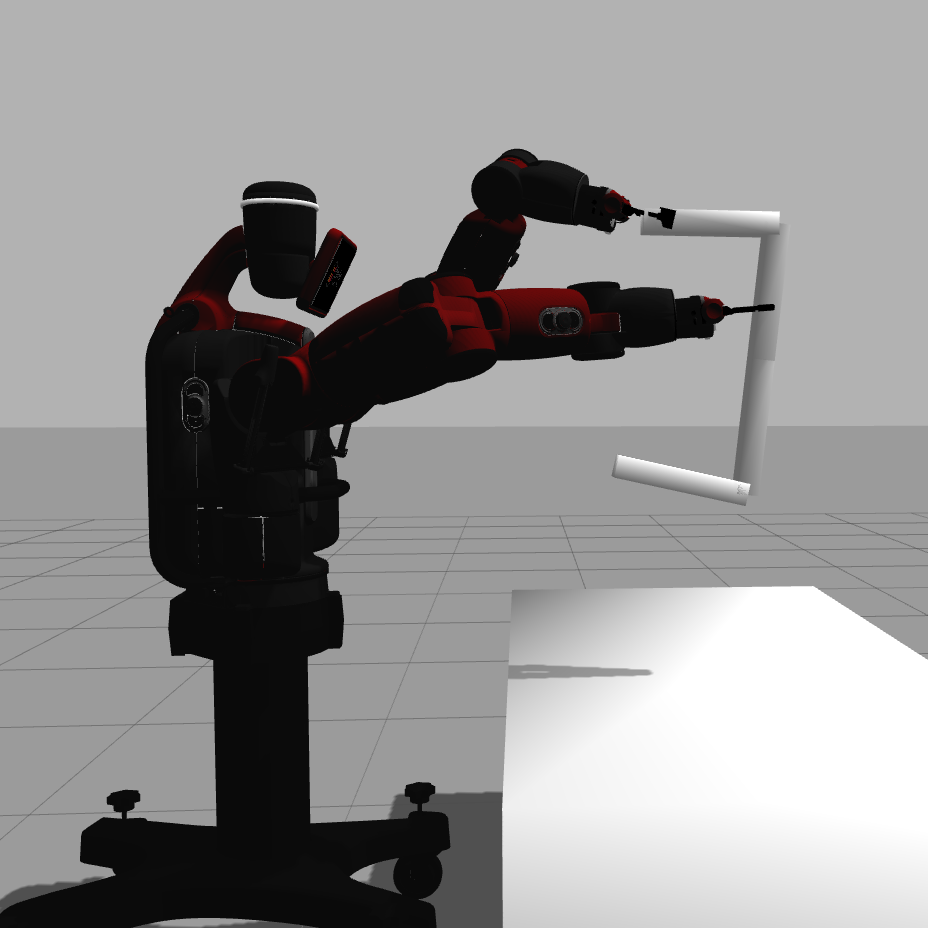}
        \label{fig:step3}
    \end{subfigure}
    \begin{subfigure}[b]{0.213\textwidth}
        \includegraphics[width=\textwidth]{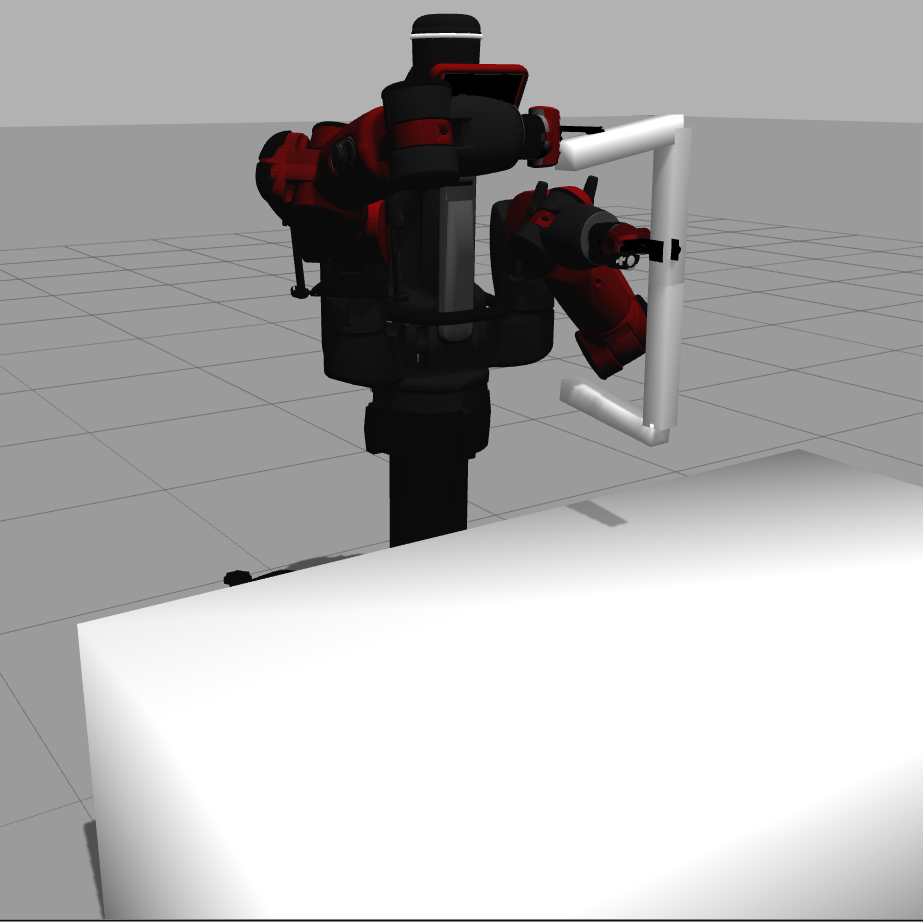}
        \label{fig:step4}
    \end{subfigure}
    \caption{Actions performed executing the sample plan described in Figure \ref{fig:plan}. From the top-left: step $1.00$, step $10.0$, step $12.0$, step $13.0$.}
    \label{fig:planexp}
\end{figure}
%Qualitatively, we observe that the motion planning and execution module is capable of successfully executing all the planned actions. 
%As a reference, Figure \ref{fig:plan} shows an example of plan employing the MCU formulation. %, as it seems to be enough to solve the tasks at hand. 
%A sample plan is reported in Figure \ref{fig:plan}, while 
%Figure \ref{fig:planexp} depicts, instead, intermediate execution steps. 
Figure \ref{fig:planexp} shows a few snapshots associated with part of the execution of the plan in Figure \ref{fig:plan}. 
Qualitatively, we observe that the motion planning and execution module is capable of successfully executing all the planned actions. 
We can also note that the UCM formulation provides an appropriate level of representation to model the effects of gravity, since generated plans can be executed in the simulated scenario without requiring adjustments or re-planning phases.

%The performed validation confirms that the plans generated by the proposed approach are appropriate to solve the task of manipulating objects in a 3D workspace. It also highlighted that the MCU formulation provides an appropriate level of representation of gravity effects, as generated plans can be executed in the simulated scenario without requiring adjustments or replanning. 

\section{Empirical Analysis: Computational Performance and Gravity}
%\subsection{}
\label{sec:cpg}

This Section presents results related to the computational performance of the planning process when different formulations of the effects of gravity are adopted.
The main aim of this analysis is to test whether our overall PDDL+ formulation can solve tasks modelling practical, real-world applications, specifically in terms of robot workspace, number of links, and physical features characterising articulated objects, and doing so in a way compatible with the timing of human-robot collaboration processes. 

We generated planning instances by varying the following parameters:
(i) the number of links $L$ of the articulated object, i.e., $3$, $4$, $5$, $6$, $7$, $8$, $10$, $12$, henceforth referred to as the object \textit{size};
(ii) in UCM: the angular speed of $0.1$, $0.5$, and $1.0$ degrees per second;
(iii) in UACM: the acceleration of $0.1$ and $0.5$ degrees per second.
In order to guarantee a fair performance assessment according to the level of complexity used for encoding the effects of gravity, for each size of the articulated object, $5$ manipulation tasks were created by randomly generating initial and final configurations.
Those instances are then encoded in PDDL+ according to the complexity level and the value of the corresponding UCM or UACM parameters.
Beside the object size, no additional parameters have to be set for the NG formulation.
Therefore, for each object size there are $5$ tasks, encoded in $30$ different problem models.
The total number of problem models considered in our experimental analysis is $240$.
Experiments have been run on a machine equipped with i7-6900K $3.20$ Ghz CPU, $32$ GB of RAM, running Ubuntu 16.04.3.LTS OS. 
$8$ GB of memory were made available for each planner run, and a $5$ CPU-time minutes cut-off time limit was enforced. 

\begin{figure}[t!]
\centering
\includegraphics[width=.45\textwidth]{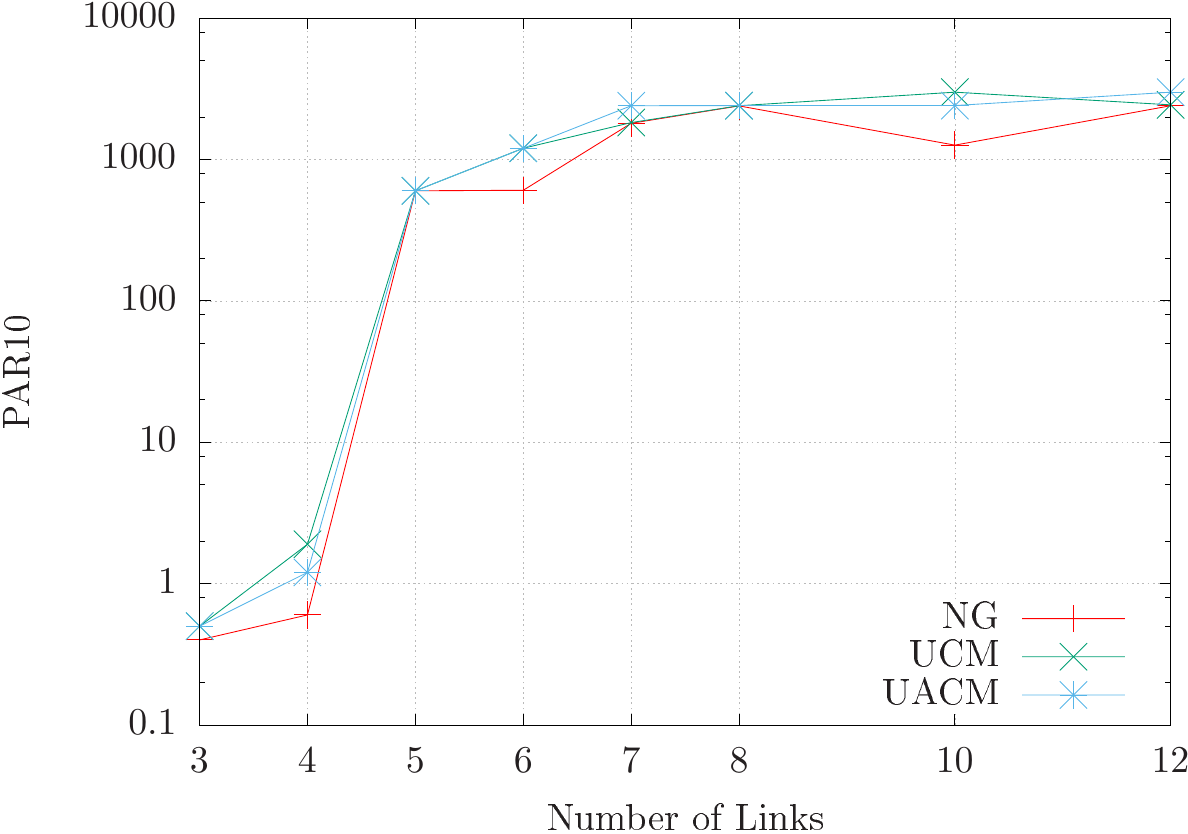}
\caption{The trend in PAR10 score achieved by ENHSP on the considered problem models, when a speed of $0.1$ degrees per second is used for UCM and UACM.}
\label{fig:comparison}
\end{figure}

\begin{table*}[t]
%\footnotesize
\centering
\begin{tabular}{ll|c|c|c|c|c|c|c|c}
 & & \multicolumn{8}{c}{Size} \\
 & & 3 & 4 & 5 & 6 & 7 & 8 & 10 & 12 \\
\hline
\hline
\textbf{NG} & & 0.4 (100) & 0.6 (100) & 0.7 (80) & 7.3 (80) & 15.5 (40) & 3.8 (20) & 108.4 (60) & 4.5 (20)   \\
\hline
\hline
\multirow{3}{*}{\bf UCM} & 0.1 & 0.5 (100) & 1.9 (100) & 1.3 (80) & 4.4 (60) & 63.9 (40) & 36.0 (20) & -- (0) & 198.5 (20)  \\
 & 0.5 & 0.5 (100) & 1.9 (100) & 1.3 (80) & 14.9 (80) & 14.5 (40) & 60.7 (20) & -- (0) & -- (0)   \\
 & 1.0 & 0.5 (100) & 1.7 (100) & 40.0 (80) & 3.2 (80) & 15.4 (40) & 55.3 (20) & -- (0) & -- (0)   \\
\hline
\hline
\multirow{2}{*}{\bf UACM} & 0.1 & 0.5 (100) & 1.2 (100) & 1.3 (80) & 4.4 (60) & 19.3 (20) & 42.5 (20) & 50.3 (20) & -- (0)   \\
 & 0.5 & 0.5 (100) & 1.2 (100) & 1.3 (80) & 4.8 (60) & 19.4 (20) & 50.0 (20) & 55.3 (20) & -- (0)  \\         
\hline
%\hline
\end{tabular}
\caption{Results achieved by ENHSP. For each articulated object's size, results are presented in terms of \textit{average run-time} (\textit{percentage of solved instances}). The average in seconds is calculated by considering solved instances only.}
\label{tab:ENHSP}
\end{table*}
Table \ref{tab:ENHSP} presents aggregated results.
The columns in the Table report results for the various sizes of the articulated object, while in the rows there are different formulations, with their variants.
For each introduced gravity encoding and size, the average run-time for solved instances is reported, while in parentheses the percentage of solved instances is indicated.
As one would expect, we observe from the results presented in the Table that the difficulty in solving the instances increases with the object size.
With regards to the level of complexity in modelling the effects of gravity, the NG model (which ignores gravity completely) seems to be the easiest (relatively) to solve, followed by UACM, while UCM seems to be the hardest.
Intuitively, the fact that UACM is easier to solve than UCM could be due to the fact that in the UACM model the effects of gravity slowly build up, while in the UCM formulation the impact of gravity starts immediately at full speed.
In other terms, the gravity has a much more impact on the search process in the UCM model rather than in UACM.
Considering the UCM and UACM formulations separately, the performance of the planning module is not significantly affected by the employed parameters for UACM, while for the UCM formulation the situation looks different.
The performance of ENHSP can significantly differ, also in the percentage of solved instances. 
Since ENHSP can solve very complicated instances (with a size up to $12$ links), taking into account gravity becomes pivotal.
Figure \ref{fig:comparison} shows the trend in PAR10 score achieved by ENHSP, when a speed of $0.1$ degrees per second is used for UCM and UACM. 
The PAR10 score is the average runtime whereby unsolved instances count as $10$ times the cut-off time, i.e., $3,000$ CPU-time seconds.

As far as the generated plans are concerned, we observe that ENHSP \textit{tends} to provide plans whereby the robot operates directly on the links that must be rotated to reach the target configuration. 

\section{Discussion}
\label{sec:discussion}

The results shown in the previous Section target the two requirements outlined in the Introduction, i.e., the generation of plans 
(i) whose actions take gravity into account, and that 
(ii) are adaptable to unpredictable behaviour of human operators during HRC processes.

The use of PDDL+ seems a very good match to that aim, but it needs to be coupled with a PDDL+ compatible planner able to avoid redundant actions that may decrease the overall efficiency of the human-robot teams, as observed by Gombolay \textit{et al} \cite{Gombolayetal2014}.
In that sense, ENHSP not only is characterised by a very good performance in terms of object size and complexity in modelling the effects of gravity, but also minimises the generation of sub-optimal action sequences that may seem odd or redundant.
%Being based on predicative knowledge, the description of each predicate or action can be easily mapped to a human-friendly communication level, even mimicking linguistic features. 
%As a consequence, these two aspects collectively enforce both the ease-of-understanding and the ease-of-explaining needs that must characterise generated plans.
As far as gravity is concerned, we developed three different models of its effects on plan generation, which are characterised by different properties in so far as accuracy and computational performance are concerned.
As discussed above, a precise evaluation of these models is not the target of the paper.
However, the results show that UCM (which constitutes a reasonable trade-off between accuracy and computational load) allows for the generation of reasonable plans, specifically in terms of actions that can be feasible for a dual-arm manipulator, in a short amount of time for a number of links up to $6$, i.e., compatible to the timing associated with a cooperation process and with a sufficient modelling precision.
Obviously enough, real-world experiments are particularly needed to support such statement, and in fact are subject of on-going work. 
It is noteworthy that generated plans encode an estimate about \textit{when} certain physical processes, for example the motion of a certain link along a vertical axis per effect of gravity, are completed. 
This can be the basis for using planning knowledge to predict future perceptions.

The use of predicative knowledge to ground perceptual information about the robot workspace, the articulated object's pose and in principle of human operator actions allow for re-planning if needed.
Considering also the sufficient computational performance, and being able to switch if necessary among different formulations characterised by different modelling precision, it is expected that the architecture can be used in real-world HRC processes.  

\section{Conclusions}

In this paper we address two related requirements for collaborative robots operating in shop-floor environments, namely 
the ability to perform complex manipulation operations, which can take into account the effects of gravity in the plan, and
doing so while retaining the capability of reacting to unpredictable behaviours by human operators.  
Adopting a use case whereby a dual-arm Baxter manipulator operates on an articulated object in a full 3D workspace, we have developed a software architecture capable of perceiving such an object, representing its configuration, planning a sequence of actions to modify it, and executing them, as an extension to previous work \cite{Capitanellietal2018,DBLP:conf/aiia/BertolucciCMMV19}.
In doing this, the planning engine makes use of models of gravity of increasing complexity, and their effects on the object configuration.
An additional significant contribution of this work is the validation in simulation of three PDDL+ models, and of the corresponding generated plans. In particular, it has been found that a good trade-off between modelling accuracy and operationality of the models is provided by the UCM formulation.

Current work is focused on a real-world implementation of the approach, and on a principled validation of the results with human operators in shop-floor environments. We are also interested in testing the models with different types of robotic manipulators, and with different manipulation tasks -- for instance considering flexible objects. Finally, we plan to assess the capabilities of planning engines (for instance ENHSP \cite{scala2013numeric}) for supporting effective monitoring and re-planning, if discrepancies between plans and real-world are observed, and to study optimisation variants of the problem, for which solutions borrowed from SAT (e.g., \cite{GiunchigliaMT02,RosaGM08,GiunchigliaMT03}) can be considered. 
%%%%%%%%%%%%%%%%%%%%%%%%%%%%%%%%%%%%%%%%%%%%%%%%%%%%%%%%%%%%%%%%%%%%%%%%%%%%%%%%

\bibliographystyle{IEEEtran} 
\bibliography{biblio}

\end{document}